\documentclass[runningheads]{llncs}
\usepackage{graphicx}
\usepackage{amsmath}
\usepackage{booktabs}
\usepackage{tabularx}
\usepackage{color}
\usepackage{multirow}
\usepackage{booktabs}
\usepackage{bbold}
\usepackage{hyperref}
\begin{document}

\newcommand \ourmethod {Meta-hallucinator}

\title{Meta-hallucinator: Towards few-shot cross-modality cardiac image segmentation}

\titlerunning{Meta-hallucinator}
%
%
\author{Ziyuan Zhao\inst{1, 2, 3} \and Fangcheng Zhou\inst{2, 4} \and
Zeng Zeng\inst{2, 3} \and Cuntai Guan\inst{1} \and S. Kevin Zhou\inst{5,6}}

%
\authorrunning{Zhao et al.}
%
\institute{Nanyang Technological University, Singapore \and
Institute for Infocomm Research (I$^2$R), A*STAR, Singapore \and
Artificial Intelligence, Analytics And Informatics (AI$^3$), A*STAR, Singapore \and
National University of Singapore, Singapore \and
Center for Medical Imaging, Robotics, Analytic Computing \& Learning (MIRACLE), School of Biomedical Engineering \& Suzhou Institute for Advanced Research, 
University of Science and Technology of China, Suzhou, China \and
Key Laboratory of Intelligent Information Processing of Chinese Academy of Sciences (CAS), Institute of Computing Technology, CAS, Beijing, China
}

\maketitle              
\begin{abstract}
Domain shift and label scarcity heavily limit deep learning applications to various medical image analysis tasks.
Unsupervised domain adaptation (UDA) techniques have recently achieved promising cross-modality medical image segmentation by transferring knowledge from a label-rich source domain to an unlabeled target domain.
However, it is also difficult to collect annotations from the source domain in many clinical applications, rendering most prior works suboptimal with the label-scarce source domain, particularly for few-shot scenarios, where only a few source labels are accessible.
To achieve efficient few-shot cross-modality segmentation, we propose a novel transformation-consistent meta-hallucination framework, meta-hallucinator, with the goal of learning to diversify data distributions and generate useful examples for enhancing cross-modality performance.
In our framework, hallucination and segmentation models are jointly trained with the gradient-based meta-learning strategy to synthesize examples that lead to good segmentation performance on the target domain.
To further facilitate data hallucination and cross-domain knowledge transfer, we develop a self-ensembling model with a hallucination-consistent property.
Our meta-hallucinator can seamlessly collaborate with the meta-segmenter for learning to hallucinate with mutual benefits from a combined view of meta-learning and self-ensembling learning. Extensive studies on MM-WHS 2017 dataset for cross-modality cardiac segmentation demonstrate that our method performs favorably against various approaches by a lot in the few-shot UDA scenario.

\end{abstract}

\section{Introduction}
Deep learning has made tremendous advancements in recent years, achieving promising performance in a wide range of medical imaging applications, such as segmentation.~\cite{long2015fully,ronneberger2015u,zhou2021review}. 
However, the clinical deployment of well-trained models to unseen domains remains a severe problem due to the distribution shifts across different imaging protocols, patient populations, and even modalities. 
While it is a simple but effective approach to fine-tune models with additional target labels for domain adaptation, this would inevitably increase annotation time and cost. 
In medical image segmentation, it is known that expert-level pixel-wise annotations are usually difficult to acquire and even infeasible for some applications. 
In this regard, considerable efforts have been devoted in unsupervised domain adaptation~(UDA), including feature/pixel-level adversarial learning~\cite{zhu2017unpaired,tzeng2017adversarial,hoffman2018cycada,chen2020unsupervised}, self-training~\cite{zou2019confidence,liu2021generative}, and disentangled representation learning~\cite{yang2019unsupervised,shin2021unsupervised,lyu20213}. 
Current UDA methods mainly focus on leveraging source labeled and target unlabeled data for domain alignment. Source annotations, however, are also not so easy to access due to expert requirements and privacy problems.
Therefore, it is essential to develop a UDA model against the low source pixel-annotation regime.
For label-efficient UDA, Zhao~\emph{et al.}~\cite{zhao2021mt} proposed an MT-UDA framework, advancing self-ensembling learning in a dual-teacher manner for enforcing dual-domain consistency. 
In MT-UDA, rich synthetic data was generated to diversify the training distributions for cross-modality medical image segmentation, thereby requiring an extra domain generation step in advance. 
In addition, images generated from independent networks have a limited potential to capture complex structural variations across domains.

On the other hand, many not-so-supervised methods, including self-supervised learning~\cite{zeng,hu2021semi},  semi-supervised learning~\cite{bai2017semi,zhao2021dsal,li2020transformation,li2021hierarchical}, and few-shot learning~\cite{ouyang2020self,tang2021recurrent} have been developed to reduce the dependence on large-scale labeled datasets for label-efficient medical image segmentation. 
However, these methods have not been extensively investigated for either extremely low labeled data regime,~\emph{e.g.}, one-shot scenarios or the 
severe domain shift phenomena,~\emph{e.g.}, cross-modality scenarios. 
Recent works suggest that atlas-based registration and augmentation techniques advance the development of few-shot segmentation~\cite{balakrishnan2019voxelmorph,zhao2019data} and pixel-level domain adaptation~\cite{lee2019image,robinson2020image}. 
By approximating styles/deformations between different images, these methods can generate the augmented images with plausible distributions to increase the training data and improve the model generalizability. 
However, image registration typically increases computational complexity, while inaccurate registrations across modalities can negatively impact follow-up segmentation performance, especially with limited annotations.
In this regard, we pose a natural question: \textit{How can we generate useful samples to quickly and reliably train a good cross-modality segmentation model with only a few source labels?} 
Recently, model-agnostic meta-learning~(``learning to learn”)~\cite{finn2017model} with the goal of improving the learning model itself via the gradient descent process is flexible and independent of any model, leading to broad applications in few-shot learning~\cite{zhang2018metagan,kiyasseh2021segmentation} and domain generalization~\cite{balaji2018metareg,liu2020shape,khandelwal2020domain}. 
Motivated by these observations, we argue that meta-learning can also enable the generator/hallucinator to ``learn to hallucinate" meaningful images and obtain better segmentation models under few-shot UDA settings. 
Therefore, we aim to build a meta-hallucinator for useful sample generation to advance model generalizability on the target domain using limited source annotations.

In this work, we propose a novel transformation-consistent meta-hallucination scheme for unsupervised domain adaptation under source label scarcity. More specifically, we introduce a meta-learning episodic training strategy to optimize both the hallucination and segmentation models by explicitly simulating structural variances and domain shifts in the training process.
Both the hallucination and segmentation models are trained concurrently in a collaborative manner to consistently improve few-shot cross-modality segmentation performance. The hallucination model generates helpful samples for segmentation, whereas the segmentation model leverages transformation-consistent constraints and segmentation objectives to facilitate the hallucination process.
We extensively investigate the proposed method with the application of cross-modality cardiac substructure segmentation using the public MM-WHS 2017 dataset. Experimental results and analysis have demonstrated the effectiveness of meta-hallucinator against domain shift and label scarcity in the few-shot UDA scenario.

\section{Method}

\begin{figure}[t]
    \centering
    \includegraphics[width = 0.9\columnwidth]{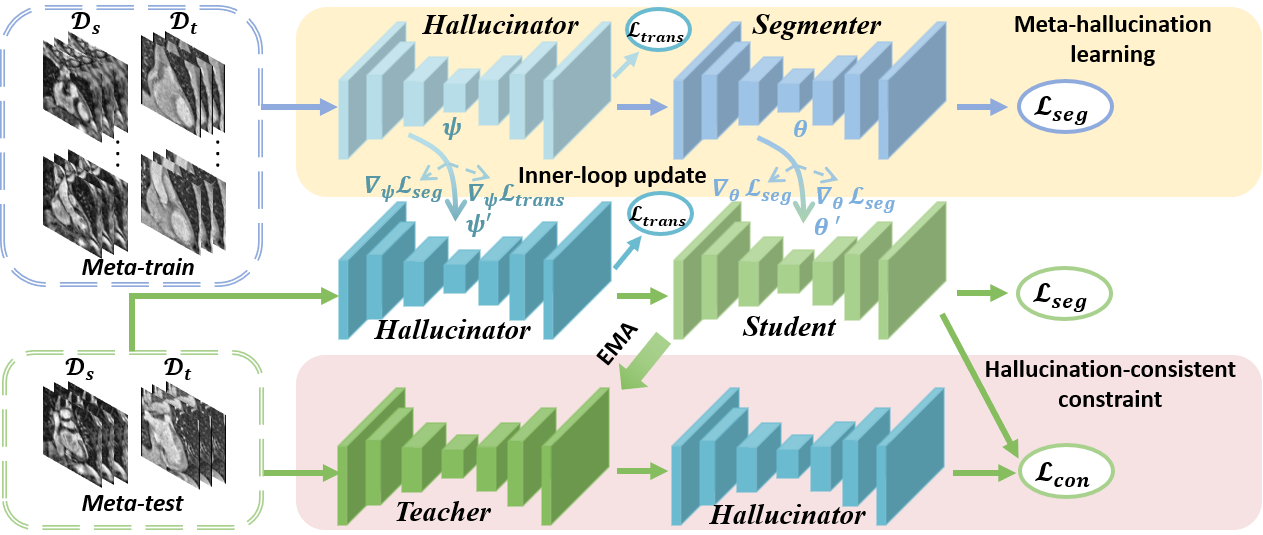}
    \caption{Overview of our transformation-consistent meta-hallucination framework. In meta-training, hallucinator $\mathcal{G}$ and segmenter $\mathcal{F}$ are optimized together with collaborative objectives. In meta-testing, the transformations generated by $\mathcal{G}$ are used for hallucination-consistent self-ensembling learning to boost cross-modality performance.}
    \label{fig:pipeline}
\end{figure}

Let there be two domains: source $\mathcal{D}_{s}$ and target $\mathcal{D}_{t}$, sharing the joint input and label space $\mathcal{X} \times \mathcal{Y}$.
Source domain contains $N$ labeled samples $\left\{\left(\mathbf{x}_{i}^{s}, y_{i}^{s}\right)\right\}_{i=1}^{N}$ and $M$ unlabeled samples $\left\{\left(\mathbf{x}_{i}^{s}\right)\right\}_{i=1}^{M}$, where $N$ is much less than $M$, while target domain includes $P$ unlabeled samples $\left\{\left(\mathbf{x}_{i}^{t}\right)\right\}_{i=1}^{P}$.
We aim to develop a segmentation model~(segmenter) $\mathcal{F}_{\theta}: \mathcal{X} \rightarrow \mathcal{Y}$ by leveraging available data and labels so that it can adapt well to the target domain.
The overview of the proposed transformation-consistent meta-hallucination framework is presented in Fig.~\ref{fig:pipeline}, which we will discuss in detail in this section.

\subsection{Gradient-based meta-hallucination learning}

In each iteration of gradient-based meta-learning~\cite{finn2017model}, the training data is randomly split into two subsets,~\emph{i.e.}, meta-train set $\mathcal{D}_{tr}$ and meta-test set $\mathcal{D}_{te}$ to simulate various tasks,~\emph{e.g.}, domain shift or few-shot scenarios, for episodic training to promote robust optimization.
Specifically, each episode includes a meta-train step and a meta-test step.
In meta-training, the gradient of a meta-train loss $\mathcal{L}_{meta-train}$ on $\mathcal{D}_{tr}$ is first back-propagated to update the model parameters $\theta \rightarrow \theta^{\prime}$.
During the meta-test stage, the resulting model $\mathcal{F}_{\theta^{\prime}}$ is further used to explore $\mathcal{D}_{te}$ via a meta-test loss $\mathcal{L}_{meta-test}$ for fast optimization towards the original parameters $\theta$. 
Intuitively, such meta-learning schemes not only learn the task on $\mathcal{D}_{tr}$, but also learn how to generalize on $\mathcal{D}_{te}$ for fast adaptation. 

In label-scarce domain shift scenarios, we are encouraged to hallucinate useful samples for diversifying training distributions to deal with label scarcity and domain shift.
To this end, we introduce a ``hallucinator" module $\mathcal{G}_{\Psi}$ to augment the training set.
The objective of the hallucinator is to narrow the domain gap at the image level and generate useful samples for boosting the segmentation performance.
We advance the hallucinator into the meta-learning process and promote it to learn how to hallucinate useful samples for the following segmentation model.
Specially, in a meta-train step, the parameters $\Psi$ and $\theta$ of the hallucinator $\mathcal{G}_{\Psi}$ and the segmenter $\mathcal{F}_{\theta}$, respectively, are updated with the meta-train set $\mathcal{D}_{tr}$ via an inner-loop update, defined as:

\begin{equation}
\begin{aligned}
\psi^{\prime} \leftarrow \psi-\alpha \nabla_{\psi} \mathcal{L}_{meta-train}\left(\psi, \theta \right);\\
\theta^{\prime} \leftarrow \theta-\alpha \nabla_{\theta} \mathcal{L}_{meta-train}\left(\psi, \theta \right),
\end{aligned}
\end{equation}
where $\alpha$ denotes the learning rate of the hyperparameters.
For the meta-train loss, the segmenter is optimized using the segmentation loss $\mathcal{L}_{seg}$ on the enlarged dataset, whereas the hallucinator objective is to minimize the transformation loss $\mathcal{L}_{trans}$ between source and target images.
It is noted that the gradient of the segmentation loss is back-propagated to both hallucinator parameters $\Psi$ and segmenter parameters $\theta$. Therefore, the total meta-train objective is defined as:

\begin{equation}
\mathcal{L}_{meta-train} = \mathcal{L}_{seg} + \lambda_{trans}\mathcal{L}_{trans},
\end{equation}
where $\lambda_{trans}$ is the weighting trade-off parameter.
For an input pair consisting of a moving source image and a fixed target image $\{x_i^s, x_i^t\}$, the hallucinator aims to generate a moved target-like image $x_i^{s \rightarrow t}$.
We promote fast and robust optimization of both hallucinator and segmenter by sampling tasks of different input pairs for meta-training and meta-testing to simulate structural variances and distribution shifts across domains.

\subsection{Hallucination-consistent self-ensembling learning}
To effectively leverage the rich knowledge hidden in the unlabeled data, we take advantage of the mean-teacher model based on self-ensembling~\cite{tarvainen2017mean}.
Specially, we construct a teacher $\mathcal{F}^{tea}$ with the same architecture as the segmenter and update it with an exponential moving average (EMA) of the segmenter parameters $\theta$ at different training steps,~\emph{i.e.}, $\theta_{t}^{tea}=\beta \theta_{t-1}^{tea}+(1-\beta) \theta_{t}$, where $t$ and $\beta$ represent the current step and the EMA smoothing rate, respectively.
With a larger $\beta$, the teacher model is less reliant on the student model parameters.
In general self-ensembling learning, the predictions of the student and teacher models with inputs under different perturbations, such as noises are encouraged to be consistent for model regularization,~\emph{i.e.}, $\mathcal{F}^{tea}\left(x_{i} ; \theta_t^{tea}, \xi^{\prime}\right) = \mathcal{F}\left(x_{i} ; \theta_t, \xi\right)$, where $\xi^{\prime}$ and $\xi$ represent different perturbations.
In contrast to the geometric transformation-invariant property in the context of classification tasks, segmentation is desired to be transformation equivariant at the spatial level. In other words, if the input is transformed with a function $f$, the output should be transformed in the same manner.
Several previous studies~\cite{li2020transformation,zhao2020sess} have demonstrated that the transformation consistency is beneficial for enhancing the regularization of self-ensembling models via various transformation operations, such as rotation. 
In light of these, we introduce a hallucination-consistent self-ensembling scheme to further promote unsupervised regularization. 
We apply the same spatial transformations produced by the hallucinator to the student inputs and the teacher outputs, and enable the alignment between their final outputs,~\emph{i.e.}, $\mathcal{G}_{\Psi}(\mathcal{F}^{tea}\left(x_{i}; \theta_t^{tea}\right)) = \mathcal{F}\left(\mathcal{G}_{\Psi}(x_{i}) ; \theta_t\right) $. The student model is regularized by minimizing the difference between the outputs of the student and teacher models with a mean square error~(MSE) loss. Then, the hallucination-consistent loss is defined as:
\begin{equation}
\label{eq:loss_kd}
\mathcal{L}_{con}=\frac{1}{N} \sum_{i=1}^{N}\left\|\mathcal{G}_{\Psi}((\mathcal{F}^{tea}\left(x_{i} ; \theta_t^{tea}, \xi^{\prime}\right))-\mathcal{F}\left(\mathcal{G}_{\Psi}(x_{i}) ; \theta_t, \xi\right))\right\|^{2},
\end{equation}
where $N$ denotes the number of samples. 
Different from stochastic transformations, such as random rotation, our hallucination process is learned via meta-learning, producing more meaningful target-like samples in spatial and appearance for domain adaptation. 
In addition, the hallucination consistency can be used to regularize the meta-optimization of the hallucinator. 
Note that we only impose the hallucination-consistent loss in the meta-test step since we expect such regularization on unseen data for robust adaptation, thereby improving the network generalization capacity. 
Then, the meta-test loss is defined as:
\begin{equation}
\mathcal{L}_{meta-test} = \mathcal{L}_{seg} + + \lambda_{con} \mathcal{L}_{con} + \lambda_{trans}\mathcal{L}_{trans}, 
\end{equation}
where $\lambda_{con}$ is to control the strength of the unsupervised consistency loss. Finally, the total objective of meta-learning is defined as:

\begin{equation}
\underset{\psi, \theta}{\operatorname{argmin}}~\mathcal{L}_{\text {meta-train }}\left(\mathcal{D}_{\text {tr }} ; \psi, \theta\right)+\mathcal{L}_{\text {meta-test }}\left(\mathcal{D}_{\text {te }} ; \psi^{\prime}, \theta^{\prime}\right).
\end{equation}

\section{Experiments and Results}

\noindent\textbf{Dataset and evaluation metrics.}
In light of our emphasis on cross-modality segmentation with distinct distribution shifts, we employ the public available Multi-Modality Whole Heart Segmentation (MM-WHS) 2017 dataset~\cite{Zhuang2016Multi} to evaluate our meta-hallucinator framework. 
The dataset contains unpaired $20$ MR and $20$ CT scans with segmentation maps corresponding to different cardiac substructures. For unsupervised domain adaptation, MR and CT are employed as source $\mathcal{D}_{s}$ and target $\mathcal{D}_{t}$, respectively.
Following~\cite{chen2020unsupervised},  the volumes in each domain are randomly divided into a training set ($16$ scans) and a testing set ($4$  scans).
For the study on label-scarce scenarios, experiments are conducted with $1$-shot and $4$-shots source labels. 
We repeat $4$ times with different samples for one-shot scenarios to avoid randomness. 
For pre-processing, each volume is resampled with unit spacing, and the slices centered on the heart region in the coronal view are cropped to $256 \times 256$ and then normalized with $z$-score into zero mean and unit standard deviation. 
Four substructures of interest are used for evaluation,~\emph{i.e.}, ascending aorta (AA), left atrium blood cavity (LAC), left ventricle blood cavity (LVC), and myocardium of the left ventricle (MYO). 
Two commonly used metrics for segmentation,~\emph{i.e.}, Dice score~(Dice) and Average Surface Distance (ASD)~\cite{zhao2021mt} are employed to evaluate different methods. 
Both metrics are reported with the mean performance and the cross-subject variations.

\noindent\textbf{Implementation details.}
We employ the 2D U-Net~\cite{ronneberger2015u} as the segmentation model due to the large variation on slice thickness cross domains. 
For the hallucinator, we consider image-and-spatial transformer networks~(ISTNs)~\cite{robinson2020image}, including a CycleGAN-like model~\cite{zhu2017unpaired} for style translation and a spatial transformer network for image registration. Considering memory limitations, we only involve the spatial transformer network in the meta-learning process. More Specifically, we first follow CycleGAN~\cite{zhu2017unpaired} to achieve unpaired image translation for image adaptation.
Since limited labels are provided in the source domain, we transform target images to source-like images for training and testing.
Then, the pairs of source images and source-like images are fed into our scheme for augmentation and segmentation. 
For segmentation loss, we use the combination of Dice loss and Cross-entropy loss~\cite{zhao2021mt}, while the transformation loss involved in meta-learning is based on MSE loss between source images and source-like images~\cite{robinson2020image}.
We train the whole framework for $150$ epochs using Adam optimizer. The batch size is set as $32$, including $8$ labeled data, $8$ augmented data, and $16$ unlabeled data. The number of pairs for meta-train and meta-test are set as $16$ and $8$, respectively. The learning rate for inner-loop update is set as $0.001$. The learning rate for meta-optimization is linearly warmed up from $0$ to $0.005$ over the first $30$ epochs to avoid volatility in early iterations. The EMA decay rate $\beta$ is set as $0.99$, and hyperparameters $\lambda_{con}$ and $\lambda_{trans}$ are ramped up individually with the function $\lambda(t)= 10\times \exp\{-5\left(1-t / 150\right)^{2}\}$ ($t$ denotes the current iteration). 
We apply data augmentations, like random rotation, and extract the largest connected component for each substructure in the 3D mesh for post-processing in all experiments.

\begin{table}[t]
\centering
\setlength\tabcolsep{4pt}
\caption{Segmentation performance of different approaches.}
\label{tab:results}
\scalebox{0.7}{
\begin{tabular}{l|ccccc|ccccc} 
\toprule
\multirow{2}{*}{Method} & \multicolumn{5}{c|}{Dice (\%)~$\uparrow$}                                                                                                      & \multicolumn{5}{c}{ASD (voxel)~$\downarrow$}                                                                                                     \\ 
\cline{2-11}
                        & \multicolumn{1}{c}{AA} & \multicolumn{1}{c}{LAC} & \multicolumn{1}{c}{LVC} & \multicolumn{1}{c}{MYO} & \multicolumn{1}{c|}{Average} & \multicolumn{1}{c}{AA} & \multicolumn{1}{c}{LAC} & \multicolumn{1}{c}{LVC} & \multicolumn{1}{c}{MYO} & \multicolumn{1}{c}{Average}  \\ 
\hline
\multicolumn{11}{c}{4-shots}                                                                                                                                                                  \\ 
\hline
Supervised-only                                                   & 85.0$_{9.2}$  & 87.1$_{3.7}$  & 75.2$_{11.6}$ & 63.0$_{20.2}$ & 77.6$_{11.2}$    & 2.2$_{0.7}$   & 3.3$_{1.3}$   & 5.1$_{3.0}$   & 5.1$_{3.8}$   & 3.9$_{2.2}$     \\
W/o adaptation                                                    & 18.9$_{19.7}$ & 4.6$_{4.6}$   & 20.7$_{17.0}$ & 11.6$_{12.5}$ & 14.0$_{13.5}$    & 48.2$_{34.0}$ & 30.8$_{15.0}$ & 35.2$_{12.7}$ & 40.6$_{22.6}$ & 38.7$_{21.1}$     \\ 
\hline
ADDA~\cite{tzeng2017adversarial}                                                             & 35.5$_{24.3}$ & 4.2$_{4.2}$   & 2.1$_{3.6}$   & 36.9$_{3.9}$  & 19.7$_{9.0}$    & 30.9$_{38.0}$ & 47.9$_{31.2}$ & 57.4$_{21.7}$ & 11.8$_{3.5}$  & 37$_{23.6}$       \\
CycleGAN~\cite{zhu2017unpaired}                                                         & 43.7$_{27.7}$ & 49.8$_{12.2}$ & 43.2$_{26.9}$ & 23.1$_{24.7}$ & 40.0$_{22.9}$      & 25.4$_{18.3}$ & 12.9$_{4.7}$  & 17.9$_{19.0}$ & 36.4$_{31.5}$ & 23.1$_{18.4}$      \\
SIFA~\cite{chen2020unsupervised}                                                             & 42.3$_{17.4}$ & 61.0$_{6.6}$  & 46.4$_{21.1}$ & 42.0$_{20.2}$ & 47.9$_{16.3}$    & 10.2$_{3.4}$  & 8.0$_{2.8}$   & 10.5$_{5.6}$  & 8.2$_{4.3}$   & 9.2$_{4.0}$      \\
\hline
MT~\cite{tarvainen2017mean}                                                                & 59.0$_{1.8}$  & 59.3$_{25.8}$ & 45.3$_{19.2}$ & 35.9$_{19.2}$ & 49.9$_{16.5}$    & 6.8$_{1.0}$   & 6.6$_{3.6}$   & 10.3$_{6.8}$  & 9.9$_{6.2}$   & 8.4$_{4.4}$       \\
TCSM~\cite{li2020transformation}                                                              & 65.3$_{3.1}$ & 62.7$_{16.9}$ & 50.9$_{13.0}$ & 38.3$_{7.8}$ & 54.3$_{10.2}$    & 5.6$_{0.6}$   & 6.2$_{2.3}$   & 9.6$_{6.6}$   & 8.2$_{2.8}$  & 7.4$_{3.1}$      \\ 
ISTN~\cite{robinson2020image}                                                              & 34.0$_{12.0}$ & 61.0$_{14.6}$ & 47.1$_{17.3}$ & 32.9$_{13.6}$ & 43.8$_{14.4}$    & 10.0$_{1.6}$  & 5.4$_{1.3}$   & 9.7$_{4.2}$   & 10.7$_{4.2}$  & 9.0$_{2.8}$         \\
VoxelMorph~\cite{balakrishnan2019voxelmorph}                                                        & 57.6$_{7.2}$  & 67.2$_{12.9}$ & 41.1$_{21.0}$ & 35.7$_{9.3}$  & 50.4$_{12.6}$    & 6.6$_{1.1}$   & 7.2$_{2.7}$   & 9.9$_{6.3}$   & 9.8$_{2.8}$   & 8.4$_{3.2}$       \\ 
\hline
MT-UDA~\cite{zhao2021mt}                                                           & 67.2$_{6.6}$  & $\mathbf{80.0_{4.1}}$  & 72.1$_{8.4}$  & 56.2$_{11.8}$ & 68.9$_{7.8}$     & 6.3$_{2.5}$   & $\mathbf{4.1_{1.0}}$   & 5.7$_{2.6}$   & 6.8$_{2.5}$   & 5.7$_{2.2}$       \\ 
\hline
Ours                                                              & $\mathbf{75.6_{8.3}}$  & 75.1$_{11.6}$  & $\mathbf{82.3_{4.6}}$ & $\mathbf{69.6_{6.8}}$ & $\mathbf{75.6_{11.3}}$     & $\mathbf{4.8_{2.9}}$   & $5.1_{2.5}$   & $\mathbf{4.3_{1.7}}$   & $\mathbf{4.9_{0.9}}$   & $\mathbf{4.8_{2.0}}$       \\ 
\hline\hline
\multicolumn{11}{c}{1-shot}                                                                                                                                                                 \\ 
\hline
ADDA~\cite{tzeng2017adversarial}                                                                & 17.3$_{12.4}$  & 12.7$_{7.1}$   & 15.7$_{12.1}$  & 15.2$_{11.7}$  & 15.2$_{10.8}$ & 47.1 $_{12.8}$ & 34.5$_{4.7}$   & 40.5$_{12.0}$  & 37.3$_{10.4}$  & 39.9$_{10.0}$  \\
CycleGAN~\cite{zhu2017unpaired}                                                           & 8.9$_{6.3}$    & 10.0$_{11.7}$  & 14.2$_{13.2}$  & 7.1$_{6.9}$    & 10.1$_{9.5}$  & 28.5$_{2.8}$   & 31.7$_{7.1}$   & 22.0$_{6.7}$   & 21.7$_{8.7}$   & 26.0$_{6.3}$  \\
SIFA~\cite{chen2020unsupervised}                                                                & 15.3$_{12.0}$  & 26.3$_{21.5}$  & 16.8$_{12.5}$  & 13.0$_{10.3}$  & 17.9$_{14.0}$ & 37.6$_{15.4}$  & 25.3$_{21.4}$  & 21.7$_{12.7}$  & 18.5$_{9.3}$   & 25.8$_{14.7}$   \\ 
\hline
MT~\cite{tarvainen2017mean}                                                           & 20.1$_{16.2}$  & 18.2$_{9.4}$   & 24.1$_{13.9}$  & 21.1$_{5.5}$   & 21.0$_{11.3}$ & 41.8$_{22.0}$   & 25.7$_{6.0}$   & 24.5$_{9.5}$   & 26$_{6.8}$     & 29.5$_{11.1}$  \\
TCSM~\cite{li2020transformation}                                                            & 32.7$_{15.8}$  & 30.8$_{9.3}$   & 37.7$_{13.9}$  & 20.1$_{5.3}$   & 30.3$_{11.0}$  & 28.0$_{11.2}$   & 31.9$_{12.2}$  & 23.3$_{11.0}$   & 23.1$_{7.1}$   & 26.6$_{9.0}$    \\ 
\hline
ISTN~\cite{robinson2020image}                                                           & 24.5$_{10.0}$  & 21.5$_{5.9}$   & 26.6$_{15.3}$  & 18.5$_{11.7}$  & 22.8$_{10.7}$  & 32.2$_{5.8}$    & 46.8$_{12.4}$  & 25.6$_{10.6}$  & 27.8$_{8.5}$   & 33.1$_{9.3}$    \\
VoxelMorph~\cite{balakrishnan2019voxelmorph}                                                      & 18.9$_{6.1}$   & 25.7$_{5.8}$   & 28.6$_{11.3}$  & 23.4$_{7.6}$   & 24.2$_{7.7}$   & 45.9$_{9.4}$    & 28.8$_{4.9}$   & 21.8$_{5.0}$   & 21.8$_{5.2}$   & 29.6$_{6.1}$    \\ 
\hline
MT-UDA~\cite{zhao2021mt}                                                           & 37.6$_{11.3}$  & $\mathbf{43.6_{11.1}}$  & 47.5$_{15.2}$  & 36.0$_{5.7}$   & 41.2$_{10.8}$  & 26.8$_{11.4}$   & $\mathbf{23.5_{12.2}}$  & 16.8$_{4.7}$   & 16.7$_{3.0}$   & 21.0$_{7.8}$    \\ 
\hline
Ours                                                              & $\mathbf{64.4_{10.3}}$  & 30.9$_{10.1}$   & $\mathbf{59.1_{6.6}}$   & $\mathbf{52.9_{5.0}}$   & $\mathbf{51.8_{8.0}}$  & $\mathbf{6.3_{1.7}}$   & 33.6$_{26.5}$   & $\mathbf{8.5_{1.7}}$     & $\mathbf{7.9_{1.7}}$    & $\mathbf{14.1_{7.9}}$   \\
\bottomrule
\end{tabular}
}
\label{tab:mmwhs}
\end{table}

\noindent\textbf{Comparisons of different methods.} We implement several well-established UDA methods,~\emph{i.e.}, a feature adaptation method~(\textbf{ADDA})~\cite{tzeng2017adversarial}, an image adaptation method~(\textbf{CycleGAN})~\cite{zhu2017unpaired}, and a synergistic image and feature adaptation method~(\textbf{SIFA})~\cite{chen2020unsupervised}, two recent popular SSL methods,~\emph{i.e.}, \textbf{MT}~\cite{tarvainen2017mean} and \textbf{TCSM}~\cite{li2020transformation}, and two representative augmentation~(Aug) methods via registration, \textbf{ISTN}~\cite{robinson2020image} and \textbf{VoxelMorph}~\cite{balakrishnan2019voxelmorph}. 
It is noted that we use CycleGAN to close the domain gap at the image level for SSL and Aug methods. 
Besides, we implement the state-of-the-art few-shot UDA method, \textbf{MT-UDA}~\cite{zhao2021mt} for comparison. 
Following previous practices~\cite{chen2020unsupervised,zhao2021mt}, we conduct experiments with the \textbf{lower ``W/o adaptation" baseline} (\emph{i.e.}, directly applying the model trained with source labels to target domain) and the \textbf{upper ``Supervised-only" baseline}~(\emph{i.e.}, training and testing on the target domain).

\begin{figure}[!thb]
    \centering
    \includegraphics[width = 0.95\columnwidth]{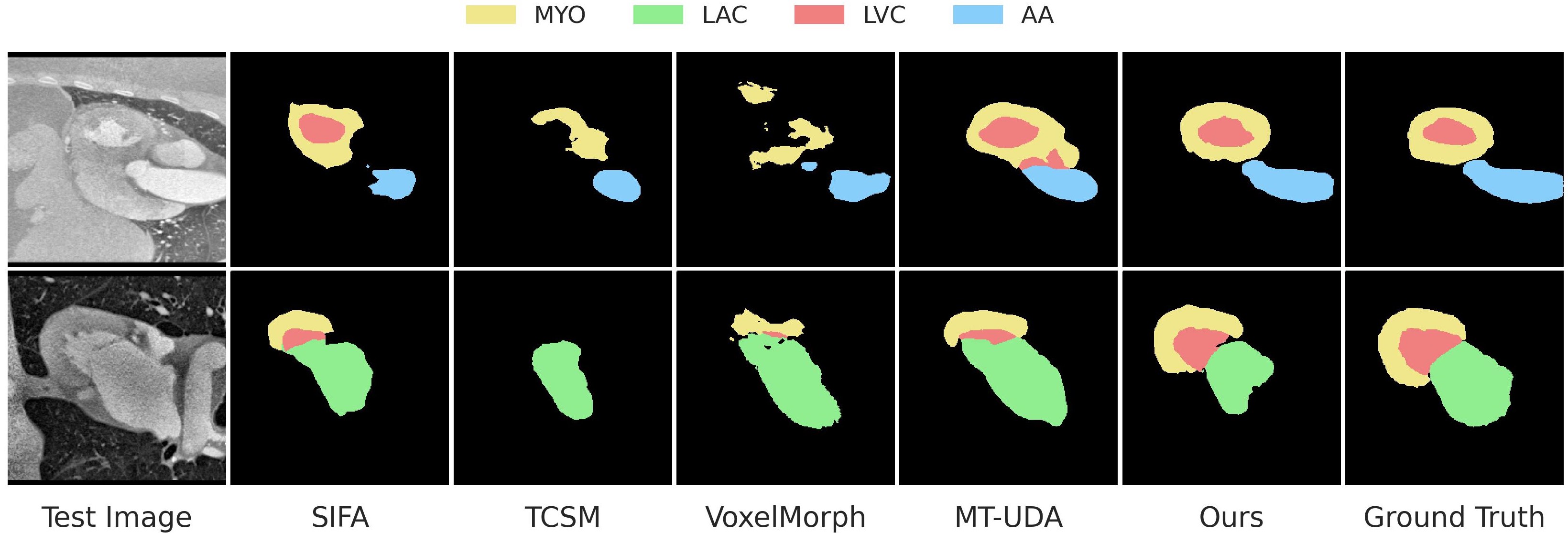}
    \caption{Visualization of segmentation results generated by different methods.}
    \label{fig:visual}
\end{figure}

The results are presented in Table~\ref{tab:results}. 
We can see that there is a significant performance gap between the upper and lower bounds due to the domain shifts. Overall, various UDA methods show unsatisfactory adaptation performance compared to the ``W/o adaptation" baseline with limited source labels.
It is observed that SSL methods,~\emph{i.e.}, MT and TCSM can help relax the dependence on source labels by leveraging unlabeled data, while Aug methods such as ISTN and VoxelMorph can also improve the segmentation performance by generating augmented samples. These results suggest that SSL and Aug methods can help unsupervised domain adaptation under source label scarcity.
Notably, our method achieves better performance than the UDA, SSL, and Aug methods by a large margin, and outperforms MT-UDA by $6.7\%$ on Dice and $0.9$mm on ASD, showing the effectiveness of our transform-consistent meta-hallucination scheme for few-shot UDA. 
With fewer source labels (1-shot), our method shows larger performance improvements than other methods, demonstrating that meta-hallucinator is beneficial in label-scarce adaptation scenarios. 
Moreover, we present the qualitative results of different methods trained on four source labels in Fig.~\ref{fig:visual} (due to page limit, we only show the best methods in UDA (SIFA), SSL (TCSM) and Aug (VoxelMorph), as well as MT-UDA. More visual comparisons are shown in Appendix). 
It is observed that our method produces fewer false positives and segments cardiac substructures with smoother boundaries.

\begin{figure}[!thb]
    \centering
    \includegraphics[width = 1\columnwidth]{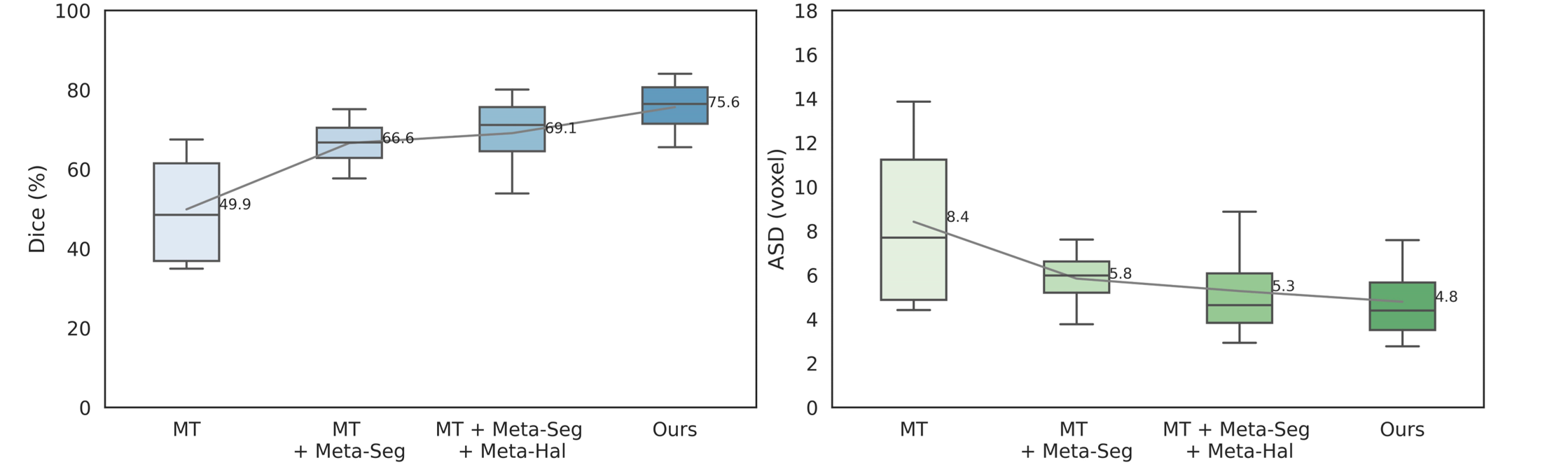}
    \caption{Boxplot of ablation results (Dice[\%] and ASD [voxel]) on different components.}
    \label{fig:abolation}
\end{figure}

\noindent\textbf{Ablation study.} Here we conduct an ablation analysis on key components of the proposed method, as shown in Fig.~\ref{fig:abolation}. 
We start by advancing mean teacher~(MT) into meta-learning with $\mathcal{L}_{Seg}$,~\emph{i.e.}, Meta-Seg, emphasizing that  Meta-Seg significantly improves the segmentation performance and outperforms most UDA methods. 
We then incorporate the hallucination module into meta-learning for data augmentation, referred to as Meta-Hal, which yields higher Dice and ASD than Meta-Seg, demonstrating the effectiveness of the meta-hallucination scheme. 
Finally, by adding hallucination-consistent constraints to enhance the regularization effects for self-ensembling training, consistent performance improvements are obtained with our method.

\section{Conclusions}
In this work, we propose a novel transformation-consistent meta-hallucination framework for improving few-shot unsupervised domain adaptation in cross-modality cardiac segmentation. 
We integrate both the hallucination and segmentation models into meta-learning for enhancing the collaboration between the hallucinator and the segmenter and generating helpful samples, thereby improving the cross-modality adaptation performance to the utmost extent. 
We further introduce the hallucination-consistent constraint to regularize self-ensembling learning simultaneously. 
Extensive experiments demonstrate the effectiveness of the proposed meta-hallucinator. 
Our meta-hallucinator can be integrated into different models in a plug-and-play manner and easily extended to various segmentation tasks suffering from domain shifts or label scarcity.

\bibliographystyle{splncs04}
\bibliography{refs}
\end{document}


\newcommand \ourmethod {Meta-hallucinator}

\title{Meta-hallucinator: Towards few-shot cross-modality cardiac image segmentation - Supplementary Material}

\titlerunning{Meta-hallucinator}
%
%
\author{Paper ID 422}
\authorrunning{F. Author et al.}
%
\institute{
}
\maketitle              
%



\begin{figure}
    \centering
    \subfloat[4-shots]{\includegraphics[width =\columnwidth]{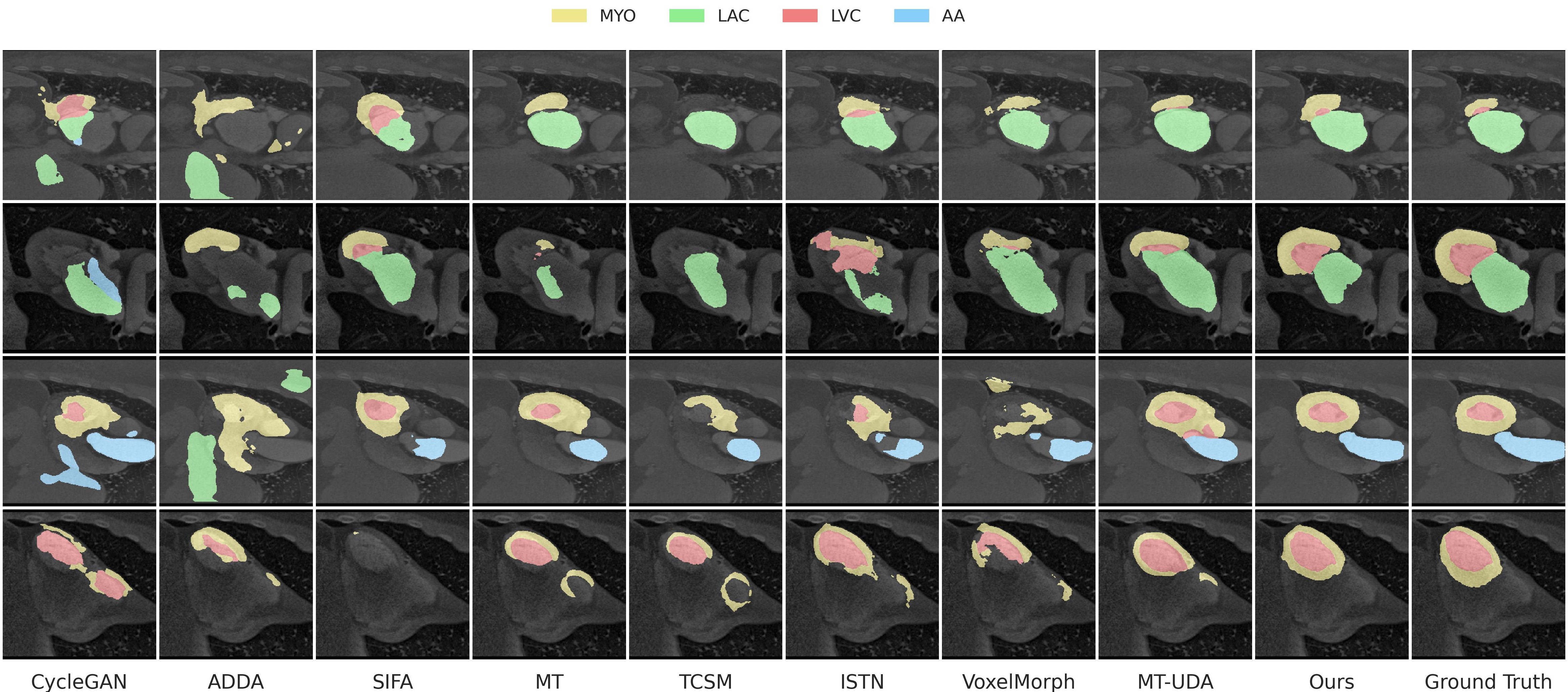}}\\
    \subfloat[1-shot]{\includegraphics[width = \columnwidth]{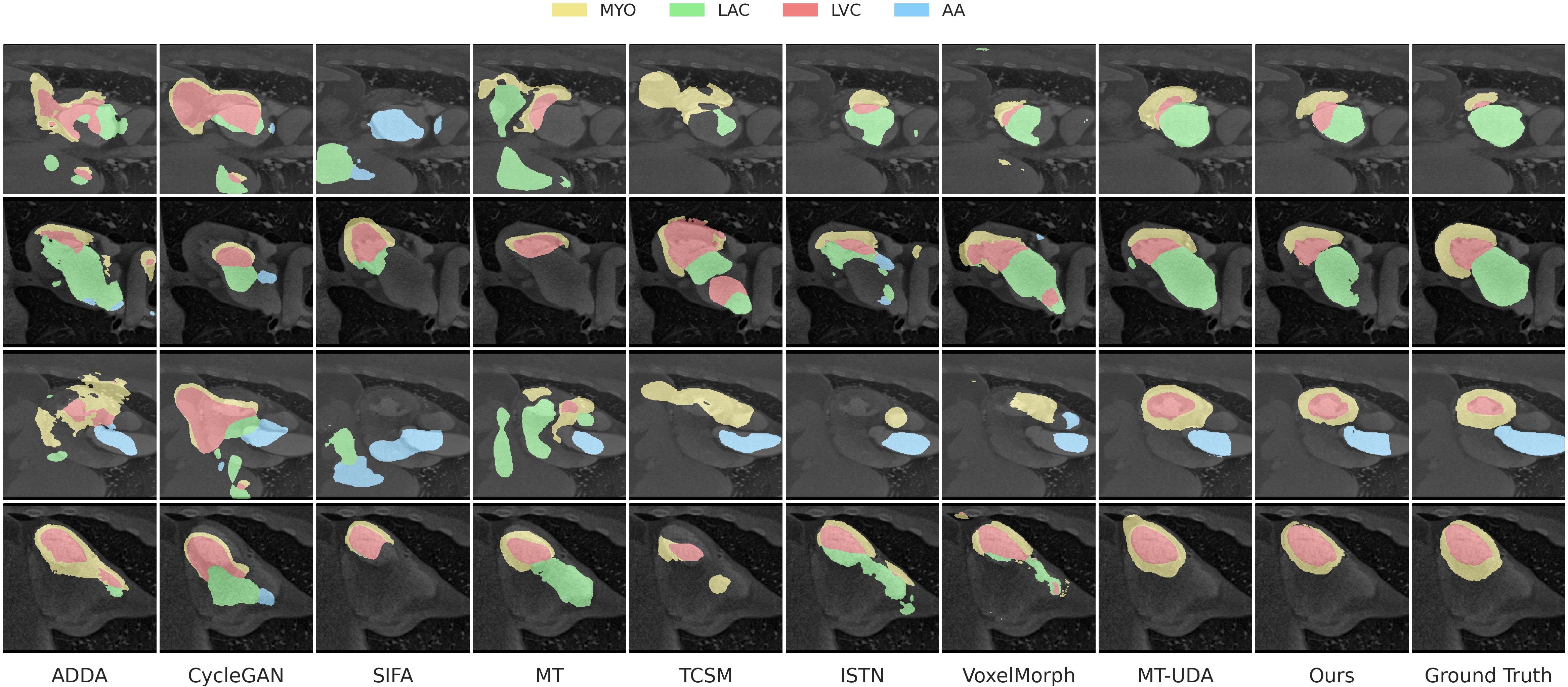}}
    \caption{Visual comparisons on MM-WHS dataset for unsupervised domain adaptation with different number of source labels.~\textbf{Best viewed in color with zoom.}}
    \label{fig:foobar}
\end{figure}

\begin{figure}[t]
    \centering
    \includegraphics[width = 0.95\columnwidth]{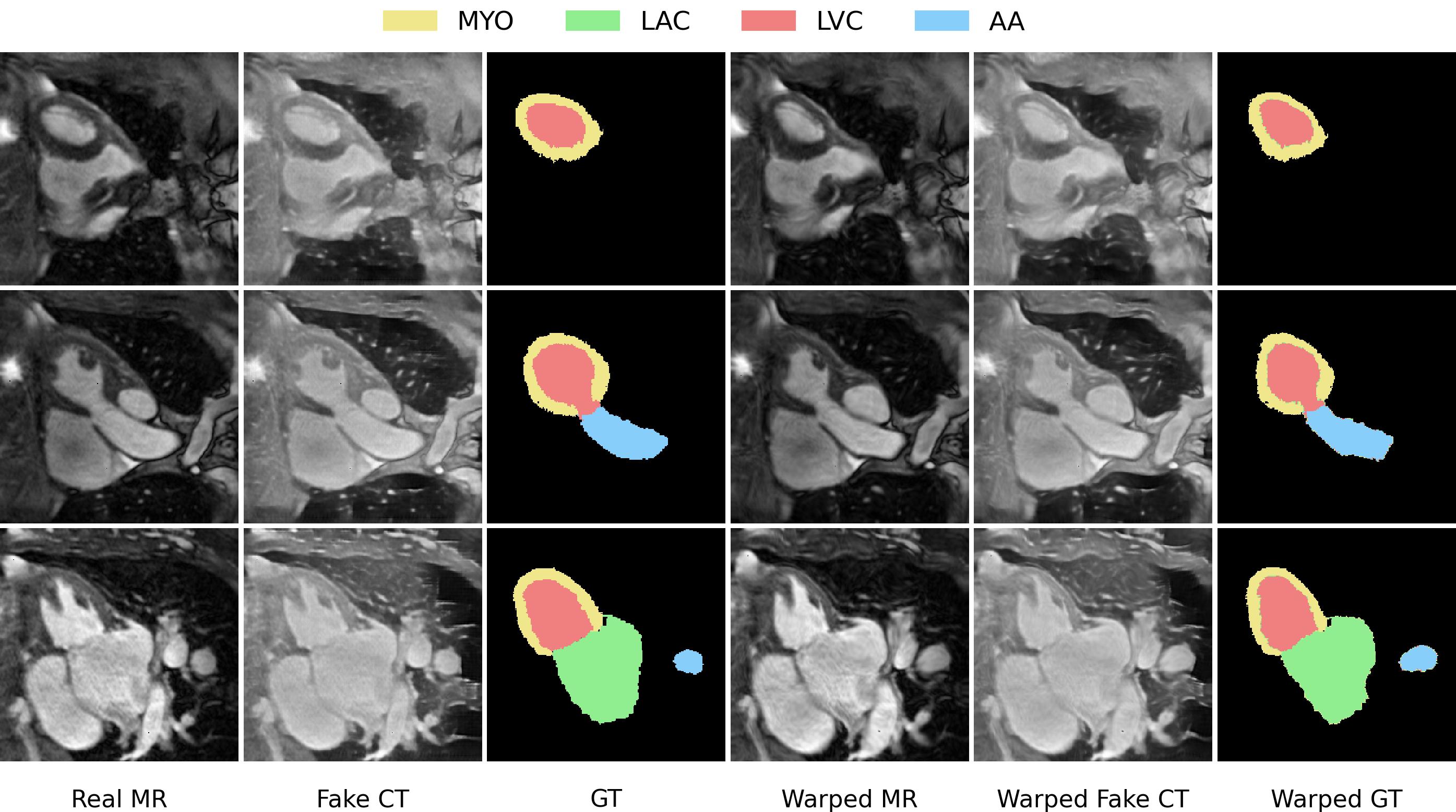}
    \caption{Visualization of some hallucinated examples. For MR-to-CT direction, the real MR images are first transformed into fake CT images with a similar appearance to CT images. Then, the obtained transformed images are warped by our method. Our warped images remain the main original contents with structural semantics while diversifying the realistic data distributions.}
    \label{fig:visual}
\end{figure}

\textbf{\begin{figure}[t]
    \centering
    \includegraphics[width = \columnwidth]{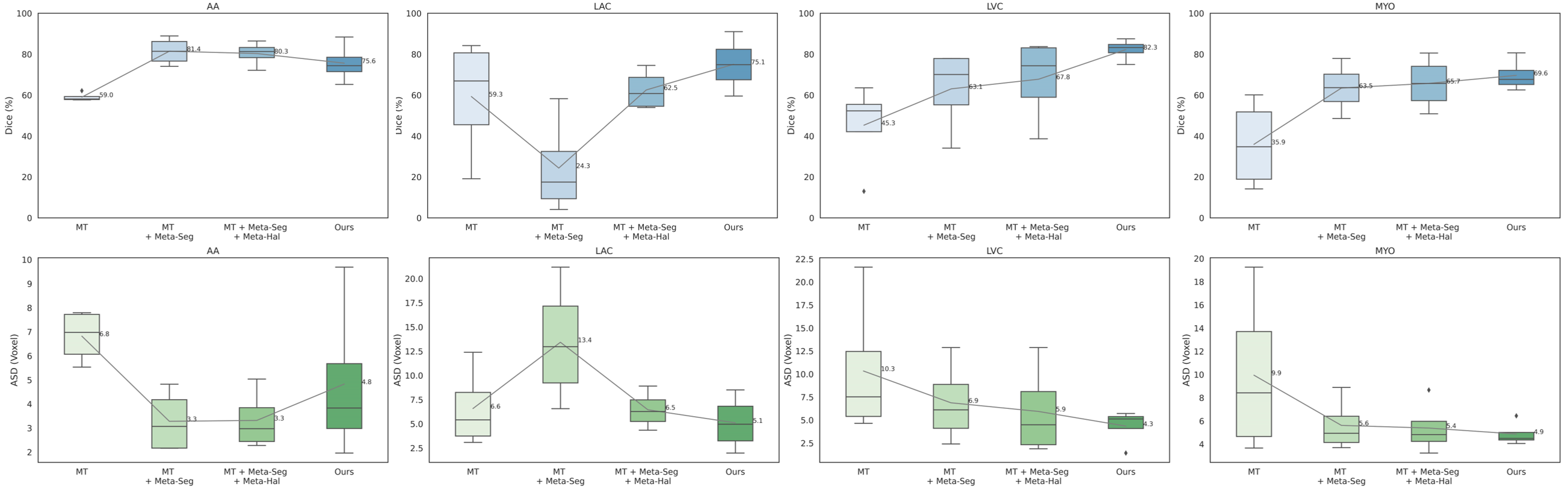}
    \caption{Boxplots of ablation results by different components in our method on four cardiac substructures.}
    \label{fig:visual}
\end{figure}}

